\documentclass[preprint,authoryear,12pt]{elsarticle}

\usepackage{amssymb,amsmath}
\usepackage{graphicx}

\usepackage{lineno}
\usepackage[usenames]{color}

\usepackage{makecell}
\usepackage{multirow}

\usepackage[hidelinks]{hyperref}
\usepackage{cite}
\usepackage{xcolor}
\usepackage{color}
\hypersetup{
    colorlinks,
    linkcolor={red!50!black},
    citecolor={blue!50!black},
    urlcolor={blue!80!black}
}

\begin{document}
\begin{frontmatter}

\title{Fully automatic detection and segmentation of abdominal aortic thrombus in post-operative CTA images using deep convolutional neural networks}
\author[add1,add2,add5]{Karen~L\'opez-Linares}
\author[add1]{Nerea~Aranjuelo}
\author[add1,add2]{Luis~Kabongo}
\author[add1,add2]{Gregory~Maclair}
\author[add1]{Nerea~Lete}
\author[add5]{Mario Ceresa}
\author[add2,add3]{Ainhoa~Garc\'ia-Familiar}
\author[add1,add2]{Iv\'an~Mac\'ia}
\author[add4,add5]{Miguel~A.~Gonz\'alez~Ballester}

\address[add1]{Vicomtech Foundation, San Sebasti\'an, Spain}
\address[add2]{Biodonostia Health Research Institute, San Sebasti\'an, Spain}
\address[add3]{Hospital Universitario Donostia, San Sebasti\'an, Spain}
\address[add4]{ICREA, Barcelona, Spain}
\address[add5]{Universitat Pompeu Fabra, Barcelona, Spain}

\begin{abstract}
Computerized Tomography Angiography (CTA) based follow-up of Abdominal Aortic Aneurysms (AAA) treated with Endovascular Aneurysm Repair (EVAR) is essential to evaluate the progress of the patient and detect complications. In this context, accurate quantification of post-operative thrombus volume is required. However, a proper evaluation is hindered by the lack of automatic, robust and reproducible thrombus segmentation algorithms. 
We propose a new fully automatic approach based on Deep Convolutional Neural Networks (DCNN) for robust and reproducible thrombus region of interest detection and subsequent fine thrombus segmentation. The DetecNet detection network is adapted to perform region of interest extraction from a complete CTA and a new segmentation network architecture, based on Fully Convolutional Networks and a Holistically-Nested Edge Detection Network, is presented. These networks are trained, validated and tested in 13 post-operative CTA volumes of different patients using a 4-fold cross-validation approach to provide more robustness to the results. Our pipeline achieves a Dice score of more than 82\% for post-operative thrombus segmentation and provides a mean relative volume difference between ground truth and automatic segmentation that lays within the experienced human observer variance without the need of human intervention in most common cases.    
\end{abstract}


\end{frontmatter}


\section{Introduction}

An Abdominal Aortic Aneurysm (AAA) is a focal dilation of the aorta that exceeds its normal diameter by more than 50\%. If not treated, it tends to grow and may rupture, with a high mortality rate~\citep{Pear08}. In the last decade, the treatment of aortic aneurysm has shifted from open surgery to a minimally invasive alternative, known as Endovascular Aneurysm Repair (EVAR)~\citep{Mol11}. This technique consists of the transfemoral insertion and deployment of a stent graft using a catheter. The prosthesis excludes the damaged aneurysm wall from blood circulation, which generates a thrombus that shrinks after the intervention in favorable cases. Despite lower rates of perioperative mortality and morbidity, studies show that two-year mortality rates are comparable to open surgery due to the appearance of EVAR complications known as endoleaks~\citep{Sta13}. These complications translate into a recurrent blood flow towards the excluded thrombus, which continues growing and needs reintervention to prevent rupture. Thus, close follow-up after EVAR is required at least yearly, for which Computed Tomography Angiography (CTA) is the preferred imaging modality~\citep{wal10}. However, this is hindered by the lack of automatic thrombus segmentation algorithms that allow precise measurement of thrombus maximum diameter, volume and other shape parameters that allow for assessment of its evolution.

Traditionally, thrombus segmentation has been addressed with intensity-based semi-automatic algorithms (level-sets, active shape models, graph cuts) combined with shape priors. Purely intensity-based techniques fail to correctly detect the non-contrasted thrombus boundaries, since there are adjacent structures that have similar intensity values into which the segmentation tends to overflow. With the insertion of a shape constraint this leakage can be further controlled. However, most of the proposed algorithms, as detailed in Sec.~\ref{sec:soa} require user interaction and/or prior lumen segmentation along with centerline extraction. Furthermore, their performance highly depends on the multiple parameter tuning, affecting the robustness and the applicability in clinical practice. We aim at exploring a new approach based on artificial intelligence that could be easily translated into clinical routine, solving some of the automation, parameter tuning, robustness, reproducibility and user interaction issues. \par

Deep Convolutional Neural Networks (DCNN) have gained attention in the scientific community for solving multiple computer vision tasks, including object recognition, classification and segmentation, surpassing the previous state-of-the-art performance in many different problems. Most importantly, DCNN methods have proven to be highly robust to varying image appearance, which is our motivation to apply them to fully automatic detection and segmentation of aortic thrombus in CTA volumes. \par

Our goal is to propose a new fully automatic approach to region of interest detection and subsequent thrombus segmentation using DCNNs. First, a 2D detection network is adapted and applied to localize the thrombus region from the complete CTA dataset. Second, we propose a new 2D DCNN architecture for the fine segmentation of the thrombus in a previously extracted region of interest. The output of the network is a probability map, so a 3D k-means based post-processing algorithm is applied to obtain the binary segmentation mask, ensuring 3D coherence and increasing accuracy while preserving low computational cost. To the best of our knowledge, this is the first study that applies deep learning to detect and segment post-operative AAA thrombi from the whole CTA volume in a fully automatic manner in a large number of images. 
The outline of the paper is as follows: Sec.~\ref{sec:soa} reviews the state of the art of thrombus segmentation approaches and DCNN for medical imaging. Sec.~\ref{sec:methods} describes our proposed method for thrombus region detection and segmentation, as well as the data employed during our experiments. Results are presented in Sec.~\ref{sec:results}. Finally, a discussion is provided in Sec.~\ref{sec:disc}.
 
\section{State-of-the-art}
\label{sec:soa}

\subsection{Thrombus segmentation}
Segmentation of the AAA thrombus is of paramount importance for diagnosis, risk assessment and determination of treatment options ~\citep{Mox10}. Segmentation of the thrombus in CTA images is challenging due to:
\begin{itemize}
\item Similarity between the intensity values of the thrombus and some adjacent tissues, causing mis-segmentation due to fuzzy borders of the thrombus
\item The thrombotic surface is locally obscured in some cases since it is a non-contrasted tissue
\item In the post-operative scenario artifacts due to the stent-graft cover some of the thrombus region
\item The geometric structure of the thrombus is irregular, which prevents the thrombus from being approximated by a simple geometrical model
\item Lack of ground truth databases
\end{itemize}

\begin{figure}[htb]
\centering
\includegraphics[width=1
\textwidth]{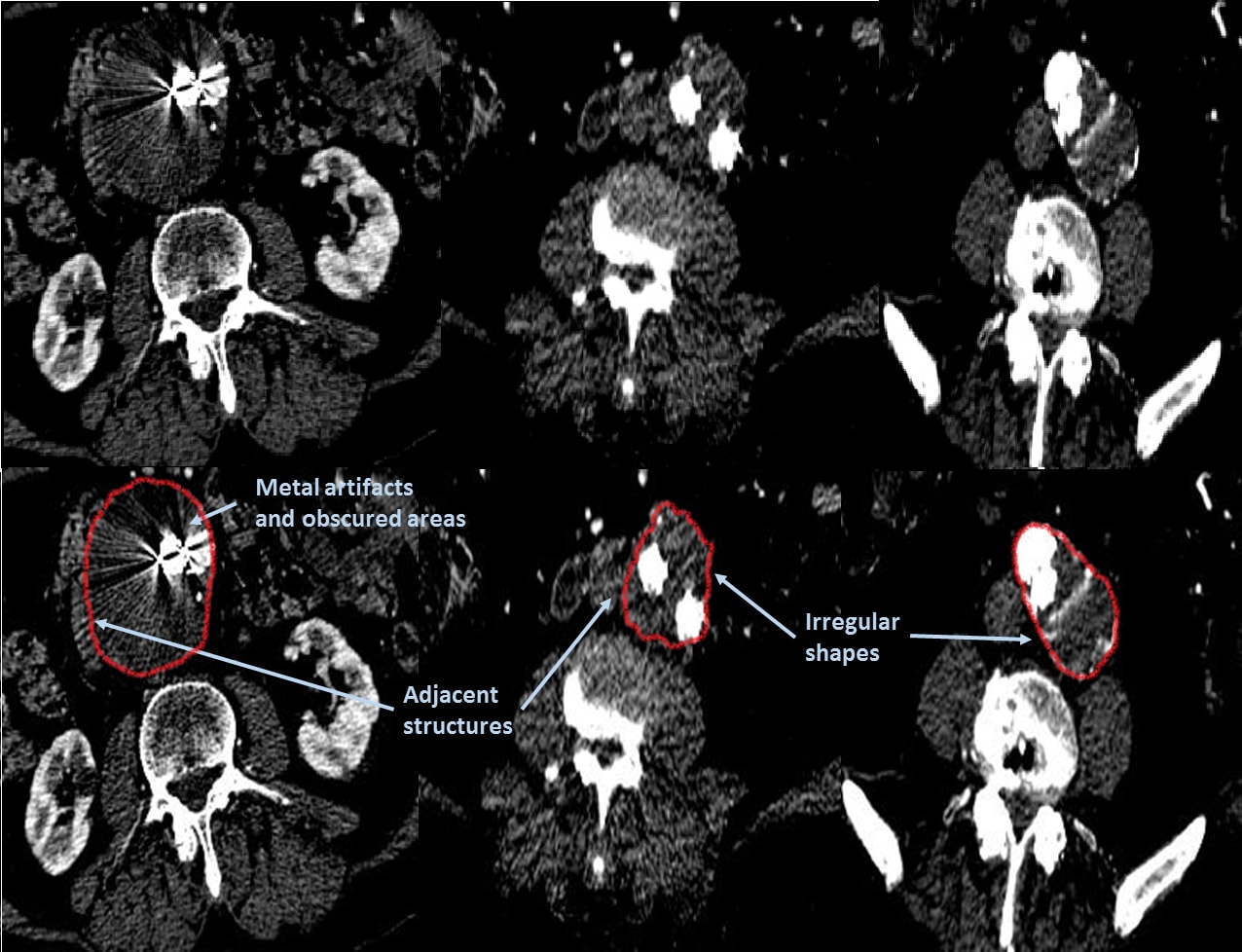}
\caption{\label{fig:problems}Segmentation training image samples depicting original images and some of the challenges when segmenting the thrombus.}
\end{figure}

These characteristics, which are shown in Figure~\ref{fig:problems}, hinder the performance of automatic thrombus segmentation approaches that are robust enough for routine follow-up. Purely intensity-based approaches fail to distinguish the thrombus from adjacent structures and thus, traditionally proposed approximations combine intensity information with shape or appearance constraints. This is the case of the approximations proposed in ~\citep{Duq12,Egg11,Mot10,kyu10}, where the authors modify the traditional 3D graph-cut energy minimization function by adding a shape model to reduce the overflow into adjacent structures. The method proposed in ~\citep{Mot10} starts from a prior segmentation of the lumen surface and consequent centerline extraction, followed by fully-automatic delimitation of the thrombus contour that yields a maximum volume overlap of 87.1\% obtained for 8 CTA datasets.  In ~\citep{Egg11}, prior extraction of the centerline is also required, as well as some manual initialization. A quantification of the volume difference between the ground truth and the obtained segmentation is not provided. The approximations proposed in ~\citep{Duq12} and ~\citep{kyu10} also rely on manual user initialization and posterior editing. A similar approximation but based in the 2D level-set equation is presented in ~\citep{Zoh12}, where several criteria are proposed to stop the evolution of the level-set curve to avoid leakage and with a strong assumption on the presence of calcifications. The method works in 2D and requires user-provided control points for the initialization. In ~\citep{dem09}, a hybrid deformable model is presented. The energy function to be minimized integrates local as well as global image information and combines it with additional shape constraints. Using cubic B-Spline surfaces as deformation models and distance functions, existing gaps in the boundary gradient are overcome and segmentation leakage into adjacent objects is prevented. The method yields good mean volume overlap measures, 93.16\%, but requires previous lumen segmentation, manual selection of some thrombus voxels and dataset-dependent parameter setup, which reduces the robustness and the reproducibility required in a real clinical setting. Radial model based approximations that assume an almost circular shape of the thrombus has also been presented in ~\citep{Mac09}.  \par
Machine learning-based approaches have also been proposed in ~\citep{Mai12, Mai14}, where thrombus segmentation is performed with active learning and supervised Random Forest (RF) classifiers. The methods are semi-automatic and no a priori geometric models are needed. In ~\citep{Mai12} the segmentation problem is addressed as a slice-by-slice multiclass classification of pixel samples. First, active learning techniques are used to select optimal feature sets, evaluating the information gain of a variety of intensity based features, to train a RF classifier and perform voxel-based segmentation, which takes about 22 minutes in total.  User interaction is needed during active learning, such that at some iterations previously misclassified data samples are added to the training set and morphological operations are required to refine the segmentation. In ~\citep{Mai14} new features for the RF classifier are included. These features are maximum, minimum, median and Gaussian weighted average of the 2D neighborhoods of the voxel of increasing radius. In both cases, classification accuracy is evaluated, but no comparison with a 3D ground truth segmentation is reported. Recently, in ~\citep{Aik16} a novel and automatic approach to pre-operative AAA region detection and segmentation is described, based on Deep Belief Networks (DBN). The detection is done in 2D, patch-wise, with patches coming from a unique dataset.  Two DBN are proposed: one detects large aneurysm patches; the other, detects small aneurysm patches, bones, organs and air. For the segmentation, another DBN is trained with 40 image patches containing aneurysm. A comparison with the ground truth is not provided. \par
From this literature review, we have determined to move towards an artificial intelligence-based segmentation approach. Hence, we propose utilizing a DCNN for CTA region of interest detection and define a novel DCNN architecture for thrombus segmentation. We also provide a complete and reproducible 3D quantitative evaluation process to compare the automatically obtained segmentations with ground truth data. In relation with previously presented approaches, our method is fully automatic and requires no parameter tuning or prior shape models. 

\subsection{Deep Convolutional Neural Networks for semantic segmentation }
In the past few years, Deep Convolutional Neural Networks (DCNN) have revolutionized the research in computer vision, since they have shown the ability to efficiently solve complex classification, segmentation, and object detection tasks. Many studies employing DCNN not only in computer vision but also in the medical imaging field have been presented, for different imaging modalities and anatomical structures, as well as network architectures.  \par
Medical image segmentation poses some inherent challenges:
\begin{itemize}
\item Different modalities have uncorrelated appearances
\item Data dimensionality is frequently higher than in 2D computer vision
\item There is little training data due to ethical concerns, data accessibility, privacy and security issues and lack of time from professionals for quality image annotation
\end{itemize}
Thus, early approaches translated the 3D medical image segmentation into a 2D patch-wise image classification approach ~\citep{Cir12,Pra13},  getting several thousands of training samples from just a few datasets. In other approaches ~\citep{Rot14} 3D data is represented as 2.5D slices, where axial, sagittal and coronal planes are stored as layers of a RGB image and the training is done with multiscale image patches.  However, according to related publications, these approximations only consider local context, are prone to failure and suffer from memory and time efficiency issues. More recent approaches combine 2D network predictions with voting approaches ~\citep{Mil16a} or traditional level-sets ~\citep{Cha16}. Spatial consistency is enforced at a second stage through post-processing computation. The natural evolution of these approximations is to train a network with the whole 2D slices. This was proposed by ~\citep{Ron15}, which defined the U-Net architecture that works with few training images and uses data augmentation strategies to compensate for the reduced amount of data and to achieve rotation or translation invariant predictions.  Fully 3D CNNs come with an increased number of parameters and significant memory and computational requirements due to 3D convolutions. In addition, the number of available medical images is always limited and training in 2D provides the ability to utilize pretrained networks and fine-tuning. We decided to leverage the advantages of 2D training and design an architecture inspired by Fully Convolutional Networks (FCN) ~\citep{Lon15} and the Holistically-nested Edge Detection (HED) network proposed in ~\citep{Xie15}. 


\subsection{Deep Convolutional Neural Networks for object detection }
In addition to segmentation, DCNN-based detection in medical imaging is also being investigated, to localize tumors, detect certain organs or structures and delimit regions of interest from complete images. Most medical detection systems based on deep learning attempt to reuse classifiers to perform the detection. These models train a classifier for a specific class or a group of classes and evaluate them at different locations and scales in the image. In ~\citep{Smi1}, in order to detect vessels in ultrasound images, some candidate regions are generated and a neural network model identifies real vessels and discards false positives from those regions. Some other methods employ a sliding window based system to run the classifier through the image. This was proposed in ~\citep{Xu1}, a breast cancer nuclei detection system that applies a sliding window operation to each testing image and feeds a classifier with those outputs to separate them as nuclear or non-nuclear. A system of these characteristics requires a quite exhaustive training and testing and has a main drawback: the final detection is based on patch classification, which lacks contextual information. Although no references to medical image detection systems trained with the whole image have been found, some computer vision approaches already proved its suitability ~\citep{Bar1,Shin1}. Recent computer vision approaches for general object detection treat the challenge as a single regression problem. In ~\citep{Red1} a neural structure that receives an input image and directly outputs bounding box coordinates and class probabilities for those boxes is proposed. The model first overlays input images with regular grids to detect if certain objects are present in those positions and then a network completes the task. In the same fashion, a network called \textit{DetectNet} is presented in ~\citep{Detectnet}. This architecture has important benefits such as optimizing the detection directly, in a one-step procedure, with no extra parameter tuning apart from the network itself. Besides, processing images once the model is trained is straightforward, without any exhaustive pipeline of candidate proposals or multiple processes. Because of its advantages we have selected this neural network to accomplish thrombus region of interest detection, translating a computer vision approach to the medical domain. This region of interest delimitation offers great benefits for the segmentation task, as it constrains the CTA region to be processed during segmentation, considerably decreasing time and memory consumption. As far as we know, no Deep Learning based abdominal aortic thrombus detection system has been proposed in the literature.

\section{Materials and methods}
\label{sec:methods}
Hereby, we propose a fully automatic pipeline for post-operative thrombus segmentation, which relies on a Deep Convolutional Neural Network (DCNN) to detect the thrombus from the whole CTA dataset followed by another DCNN for its segmentation. Deep learning approaches require a large amount of data for a good generalization and to overcome overfitting problems, but since annotated medical image data is always limited, training and testing a network directly in 3D becomes complicated. Besides, training in 2D provides advantages regarding higher speed, lower memory consumption and the ability to utilize pretrained nets and to fine-tune them. Thus, we have decided to leverage these advantages and train our networks with 2D slices, treating every CTA slice independently during training, validation and testing.

We have created our models using Nvidia Corporation's Deep Learning GPU Training System (DIGITS) based on Caffe ~\citep{caffe}. Both training and testing are done on a 2 Xeon E5620 2,4GHz, 12GB processor equipped with a Titan X graphic card donated by NVIDIA Corporation, under Linux Ubuntu 14.04 LTS 64 bits operating system.

Our method consists of two steps: 
\begin{enumerate}
\item Thrombus detection within a whole CTA volume
\item Fine thrombus segmentation from a localized Region Of Interest (ROI)
\end{enumerate}\par

The detection step consists in a DCNN network that provides 2D ROI-candidates containing the thrombus, which are then processed to get a continuous 3D region and to include some contextual information. We assume that this contextual information is necessary to obtain an accurate posterior segmentation, since relative location of the thrombus with respect to other anatomical structures may be valuable. This region of interest extraction step is essential to subsequently achieve better segmentation results, since it significantly reduces the amount of background pixels. This allows the segmentation network to have a more balanced distribution of thrombus and background pixels, improving its generalization and precision. Additionally, by reducing the size and quantity of 2D slices introduced to the segmentation network, a more computationally efficient segmentation model is obtained. 

\begin{figure}
\centering
\includegraphics[width=1
\textwidth]{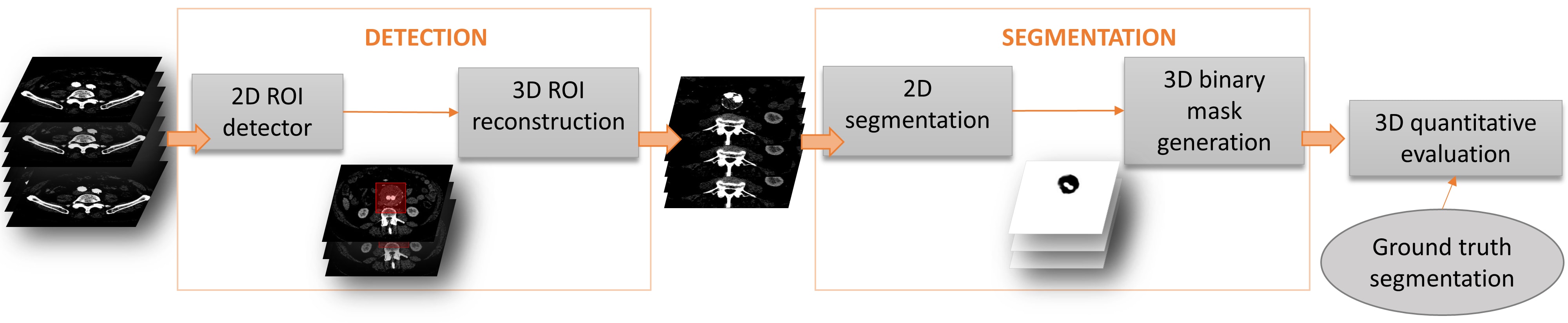}
\caption{\label{fig:method}Proposed method for thrombus segmentation.}
\end{figure}

Regarding the segmentation phase, our proposed DCNN outputs 2D probability maps that are then processed, ensuring consistency between consecutive slices, to obtain the final 3D binary segmentation. To overcome the limitations of a 2D approximation and to obtain a coherent and accurate 3D binary segmentation mask, a simple and automatic 3D K-means based post-processing step is applied to the output probability maps. Figure~\ref{fig:method} depicts a scheme of the proposed pipeline.

\subsection{Post-operative abdominal aortic aneurysm datasets}
\label{subsec:data}
The imaging data for this study consists of 13 post-operative contrast-enhanced CTA datasets from different patients. All these patients present an infrarenal aneurysm with varying evolution along time: favorable cases where the thrombus shrinks, unfavorable cases with an expanding aneurysm and a visible leak, and unfavorable endotension cases with no visible leak. As patient was excluded, the variability in the data is relatively large in terms of thrombus size and shape or noise, as shown in Figure~\ref{fig:problems}. For the CTA examination, the patient is always in supine position with the image acquisition starting around the diaphragm and expanding to the iliac crest. The datasets have been obtained with scanners from different manufacturers and models: Toshiba Aquilion, GE Lightspeed RT16, GE optima 264 CT660 and GE Lightspeed VCT. They have varying spatial resolution, ranging from 0.72-0.97 in x and y direction, and 0.625-0.8 in z direction. These datasets are employed differently for the two parts of the presented pipeline, i.e the segmentation and the detection.

For the segmentation task, manually delineated ground truth masks done by an expert vascular surgeon are available. A region of interest (ROI) is computed for each mask, which is expanded to simulate the output of the detection step that also includes contextual information. Finally, the 2D axial slices from that region of interest are extracted. 
For the training and validation of the segmentation, we have decided to directly use the images extracted from the ground truth ROIs instead of the output of the detection, to leverage all the available data and do not propagate possible detection errors (missing slices where the thrombus appears) to the segmentation. Thus, the networks are mostly trained and validated on slices where the thrombus is present, with only few slices where the thrombus is absent. Note that due to the need of contextual information, the number of pixels corresponding to the thrombus is much smaller than the number of pixels corresponding to background. For testing, slices extracted from the output of the detection network have been used to evaluate the complete pipeline.

For the detection task, all 2D axial slices of the complete CTA volumes have been extracted and the ground truths have been generated as 2D rectangular bounding boxes around the expert-outlined segmentations. In those slices where the thrombus is not visible (around 55\% of them), no bounding box is provided. 
The determination of the initial and final slices delimiting the thrombus is a complex task. From a clinical point of view, theoretically, the presence of a thrombus can be assumed when the aortic wall expands more than 5 mm. In practice, this is done approximately by visual inspection, therefore inter and intra-observer variability exists. To take this into account, we have recorded the initial and final slices where the thrombus appears in the ground truth and the slice range selected by two more experts for comparison purposes. 

\subsection{4-fold cross-validation}
Being the number of the available annotated datasets limited, a 4-fold cross-validation approach is employed to provide more robustness to our results. Each network is trained 4 times using, every time, a different (non-overlapping) set of datasets resulting in 4 different trained models or network instances. Then, each instance is tested against the remaining datasets that were excluded from training in each fold. Hence, all the 13 datasets are used for training, validating and testing the networks.

Table~\ref{tab:data} and Table~\ref{tab:dataDetect} summarize the number of 2D slices employed for each of the 4 instances of the segmentation and detection networks, respectively.   

\begin{table}[htb]
\centering
\begin{tabular}{c|c|c|c}
\hline
Fold & Training images & Validation images & Testing images \\
\hline
M\textsubscript{0} & 1519 & 168 & 501  \\
M\textsubscript{1} & 1508 & 167 & 513  \\
M\textsubscript{2} & 1498 & 166 & 524  \\
M\textsubscript{3} & 1538 & 170 & 480
\end{tabular}
\caption{\label{tab:data}Training, validation and testing data for the 4 segmentation network instances.}
\end{table}

\begin{table}[htb]
\centering
\begin{tabular}{c|c|c|c}
\hline
Fold & Training images & Validation images & Testing images \\
\hline
M\textsubscript{0} & 3323 & 359 & 1398  \\
M\textsubscript{1} & 3924 & 436 & 1044  \\
M\textsubscript{2} & 2815 & 386 & 1305  \\
M\textsubscript{3} & 3663 & 407 & 1182
\end{tabular}
\caption{\label{tab:dataDetect}Training, validation and testing data for the 4 detection network instances.}
\end{table}


\subsection{Thrombus region of interest detection}
Thrombus detection is a difficult task because of its varying size, shape and orientation. Traditionally, detection of objects with these characteristics has been addressed by manually engineered low-level feature extractors. However, these methods rely on handcrafted designs that do not learn from the data itself and are vulnerable to erroneous assumptions and lack of generalization. Deep learning methods overcome these problems due to their ability to learn more complex representations and data-driven model optimization, achieving an outstanding performance. Thus, the thrombus region detector proposed in this section is based on a trainable DCNN. The goal of this model is to detect the presence or absence of the thrombus in every 2D slice of a CTA volume and to determine a bounding box around it, then generating a continuous 3D region of interest appropriate for the segmentation.

\begin{figure}
\centering
\includegraphics[width=1
\textwidth]{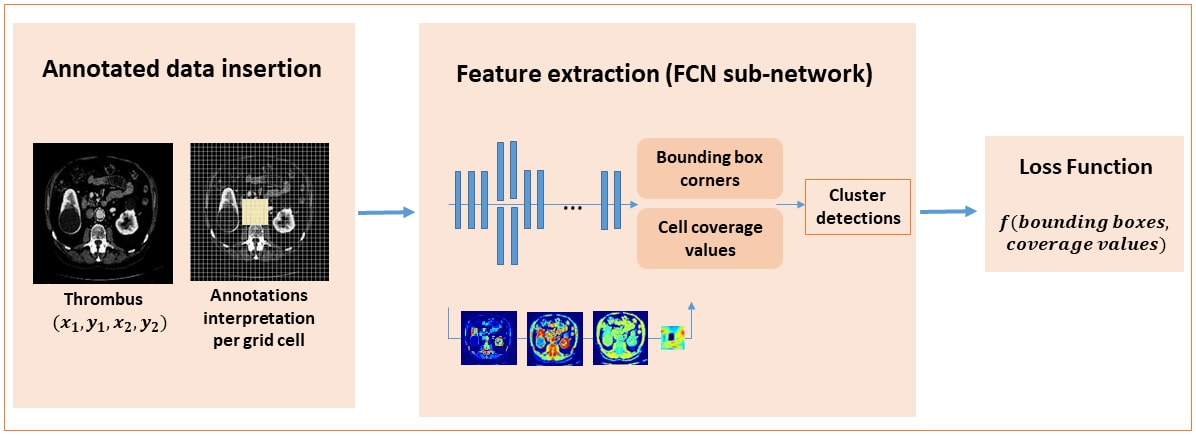}
\caption{\label{fig:detection}Thrombus detection training network scheme.}
\end{figure}

As explained in Sec.~\ref{sec:soa}, we have selected the \textit{DetectNet} ~\citep{Detectnet} neural network architecture for the thrombus detection task due to its interesting characteristics in terms of efficiency and whole image context consideration. \textit{DetectNet} simultaneously accomplishes object classification and bounding box estimation via regression. Unlike other region proposal or sliding window based methods, this architecture enables the model to analyze the entire image considering contextual information. Given an input image, the network can predict bounding boxes around the target object, i.e. the thrombus, with no need of extra processing. The network architecture consists of three modules: data insertion, core network and loss functions. A simplified scheme of the \textit{DetectNet} architecture is shown in Figure~\ref{fig:detection}.

Initially the input image is resized to 512x512 pixels and is divided into regular squared grids with constant size, in our case set to 16x16 pixels, which is the minimum size of the expected detected region. 
The purpose of each grid is to learn the probability of the thrombus being present in that cell. Each grid is labeled with a coverage value of 1 if it contains the thrombus, including also its bounding box coordinates. After that, a fully convolutional network performs feature extraction, as well as object bounding box corner prediction per grid square. This sub-network structure is based on the well-known GoogleNet ~\citep{Googlenet} architecture, with the modifications needed to convert it into fully convolutional. These adaptations mainly involve discarding the fully connected and classification layers and adding convolutional layers instead, as explained in detail in ~\citep{Lon15}. The outputs of the network are the predicted bounding box corners and the coverage values of the grid squares (1 if a thrombus is present in a certain cell and 0 if not). Before feeding the loss functions with these values, a clustering is done. The detections are filtered based on grouping rectangles with similar dimensions and locations. During training, two loss functions are considered to form the final loss function. The first term of the loss function computes the error of the object bounding box corners, that is, the mean absolute difference between true and predicted corners of the bounding boxes. The second term, measures the thrombus coverage error, computed as the sum of squares of the differences between true and predicted coverage of all cells in each training image. Both terms are linearly combined in the following minimization loss:

\[F(\textbf{c},\textbf{P})=f(\textbf{c})+f(\textbf{P})=\frac{1}{2n}\sum_{i=1}^{n}[|c_i^g-c_i^p|^2+[|P_{1i}^g-P_{1i}^p|+|P_{2i}^g-P_{2i}^p|]] \]

where \(n\) corresponds to the number of grid cells in each training image and \(c_i^g\) and \(c_i^p \) to the ground truth and predicted cell coverage values, respectively. \(P_{i1}^g\) and \(P_{i2}^g\) denote the ground truth bounding box opposite corners, represented by their 2D x and y coordinates, and \(P_{i1}^p\) and \(P_{i2}^p\) the predicted points with respect to each cell.

We have trained each detection network instance with a batch size of 10 images and ADAM optimization. We used an initial learning rate of 1e-06 and exponential learning rate decay with a gamma of 0.99. The original network proposed data augmentation inside its architecture, but we did not apply this as the augmentation did not show to improve the thrombus detection performance. All parameters were trained from scratch.


\subsubsection{3D region of interest reconstruction}
The thrombus is a continuous structure, and thus, its presence must be predicted in contiguous slices. From the output 2D bounding box candidates provided by the network, the 3D region of interest is reconstructed using a sliding window policy. From all the 2D candidates, we set the beginning of the ROI when a group of 10 contiguous slice candidates are found. The region of interest then expands until the first slice among a group of contiguous 10 slices where no bounding box candidate has been proposed. In this way, we obtain a contiguous set of slices where the thrombus presence has been detected. If more than one 3D ROI is found due to detection errors, the largest is kept, which always correspond to the thrombus, and the initial and final slice of the thrombus can be determined. The minimum and maximum x and y coordinates of the 3D region of interest are set to the corresponding values among all the 2D bounding boxes of the continuous slices and are then expanded 
to incorporate additional contextual information, leaving the thrombus somehow centered in the extracted sample. The x coordinate is expanded by around 110~mm per border to include the kidneys and the y coordinate is expanded by around 80~mm to include the spine. Finally, we rescale the size of the obtained 3D volume to 256x256x(number of slices of the ROI) to lessen the amount of memory needed during the segmentation step.

\subsection{Thrombus segmentation}
A proper thrombus segmentation approach needs to deal with the differentiation of the thrombus from adjacent structures by fine edge detection and requires that a coherent appearance of the thrombus is kept, meaning that the thrombus contour is smooth and does not present sharp edges. The first part of this section gives a brief overview of previous related segmentation networks that have served us as ground work to build up our solution in order to accomplish these two objectives.
The second part explains our proposed network for fine thrombus segmentation. We have applied cross validation in all our segmentation experiments, meaning that each network architecture is trained and tested 4 times, each of them with a different training and testing data subset. Thus, we create four network models for each architecture and average the results. This approximation provides more robustness to our approximation.
 
\subsubsection{Related networks}
\label{sec:rel_net}
\subsubsection*{Fully Convolutional Networks} 
\begin{figure}
\centering
\includegraphics[width=0.9\textwidth]{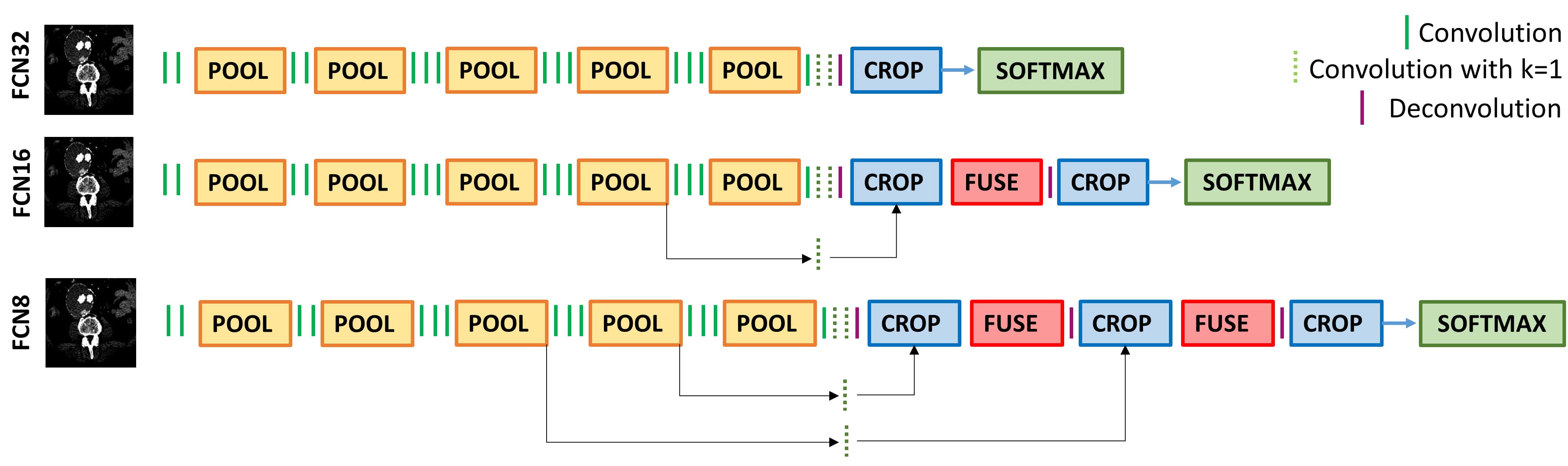}
\caption{\label{fig:fcn}Simplified scheme of FCN networks.}
\end{figure}

In 2015, researchers from UC Berkeley firstly introduced Fully Convolutional Networks (FCN) ~\citep{Lon15} for semantic segmentation by adapting classifiers for dense prediction. Semantic segmentation requires a compromise between the global information that resolves a coarse representation of the object to be segmented, and local, fine information. They showed that these networks exceed the state-of-the-art in semantic segmentation, taking an input of arbitrary size and producing a correspondingly-sized output prediction. They also introduced the skip architecture, which combines semantic information from a deep, coarse layer with appearance information from a shallow, fine layer to produce accurate and detailed segmentations. A skip-net architecture centers on a primary stream and skips or side connections are added to incorporate feature responses from different scales in a shared output layer. FCN networks with less skip connections provide more global information, while connecting more detailed feature maps with the output coarser map improves local contextualization. They presented three FCN adaptations with varying number of skip connections. Simplified schemes of these networks are shown in Figure~\ref{fig:fcn}. The networks are composed of convolution/deconvolution, pooling, crop and fuse layers. Convolutional and pooling layers subsample the input image by a certain factor or \textit{stride} value, reducing the image dimension to keep filters small and computational requirements reasonable, producing coarser outputs. To add together these varying-sized coarse outputs and to obtain the predictions upsampling is needed, which is done in the deconvolution layers. Upsampling with a factor f is convolution with a fractional input stride of 1/f. Finally, outputs of different layers must be fused, which requires alignment by scaling and cropping. Two layers are brought into scale agreement by upsampling the lower-resolution layer and cropping removes any portion of the upsampled layer which extends beyond the other layer. To guarantee that the network output can be aligned to the input for any input size without cropping relevant image information, they introduce a \textit{padding} of 100 pixels in the first convolutional layer.\par

\subsubsection*{Holistically-nested Edge Detection network}
\begin{figure}
\centering
\includegraphics[width=0.85\textwidth]{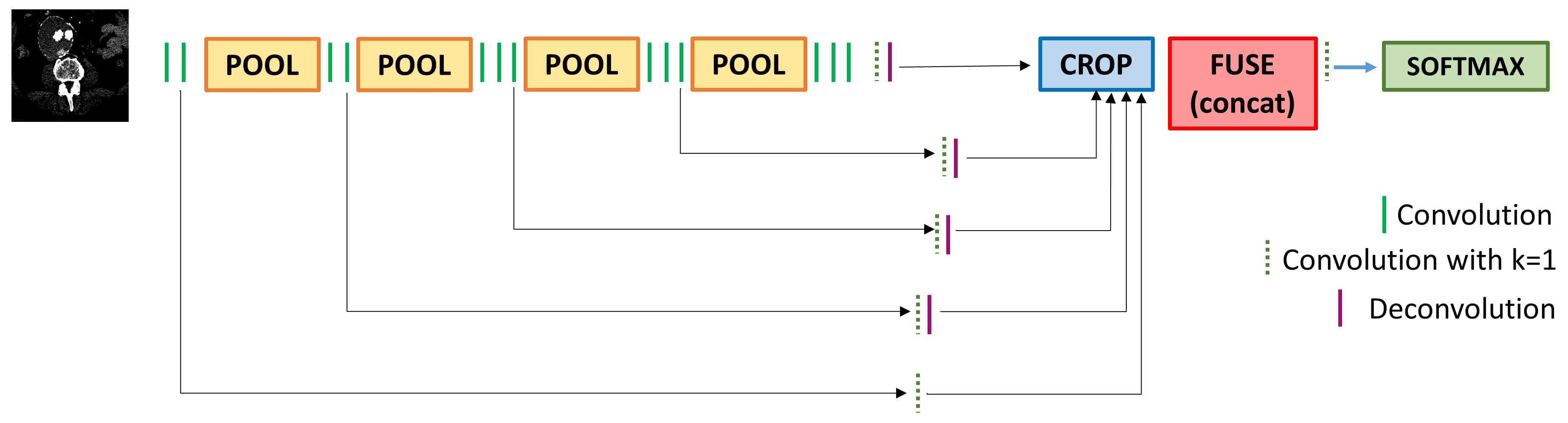}
\caption{\label{fig:hed_orig}Simplified scheme of the original HED network.}
\end{figure}

Also in 2015, researchers from the UC San Diego presented an end-to-end edge detection system, an architecture inspired by FCNs, but with additional deep supervision ~\citep{Xie15}. The network, known as Holistically-nested Edge Detection (HED) comprises a single-stream deep network with multiple side outputs. In comparison to FCN networks, they connect skip connections to the last convolutional layer in each stage, instead of after the pooling. This reduces the need of large deconvolution filters, increasing the resolution of the output prediction. They also remove the 5th pooling layer, reduce the input padding and apply a deconvolution to each side connection separately before cropping and fusing all the skip connection outputs at once. A simplified scheme of this network is shown in Figure~\ref{fig:hed_orig}.\par

\subsubsection{Proposed architecture for thrombus segmentation}
Based on the aforementioned networks, we propose a novel network architecture that detects edges but preserves the shape and appearance information of the thrombus. Our hypothesis relies on the knowledge that previous solely edge-based and intensity-based segmentation approaches fail to isolate the thrombus, due to the presence of adjacent structures with similar intensity, with no apparent edge separating it from the thrombus. The inclusion of a shape and appearance prior notably improved the results in the thrombus segmentation problem. We want to exploit this knowledge by proposing an architecture that combines different relevant scales: low scale for fine edge and border detection; and more global appearance and localization information, obtained with a macro scale, that ensures the smooth contour of the thrombus, which is located above the vertebrae, around the lumen and between both kidneys.


To compare the proposed network performance with the FCN and HED networks, we have started by adapting and fine tuning the 3 FCN networks with different skips connections described in Sec. \ref{sec:rel_net}. We have reduced the initial padding from 100 pixels to 35, accordingly increasing the stride in the last deconvolution layer: a smaller padding demands a bigger stride, which reduces the resolution of the output segmentation. However, in our case it is necessary since otherwise the number of thrombus pixels compared to the background would be hugely imbalanced. We have renamed the networks as FCN14, FCN26 and FCN46, in relation to the new stride value of the last deconvolution layer. FCN14 has no skip connection, while, FCN26 and FCN46 have 1 and 2 skip connections, respectively.
As expected from previous experiments related to the topic, the FCN14 and FCN26 networks fail to understand the global shape and appearance of the thrombus, while FCN46 is unable to differ the thrombus from adjacent, touching structures but preserves the global shape of the object and yields better results. The FCN14 and FCN26 networks try to locate fine edges, but the long stride limits the scale of detail in the upsampled output. 

\begin{figure}[htb]
\centering
\includegraphics[width=1
\textwidth]{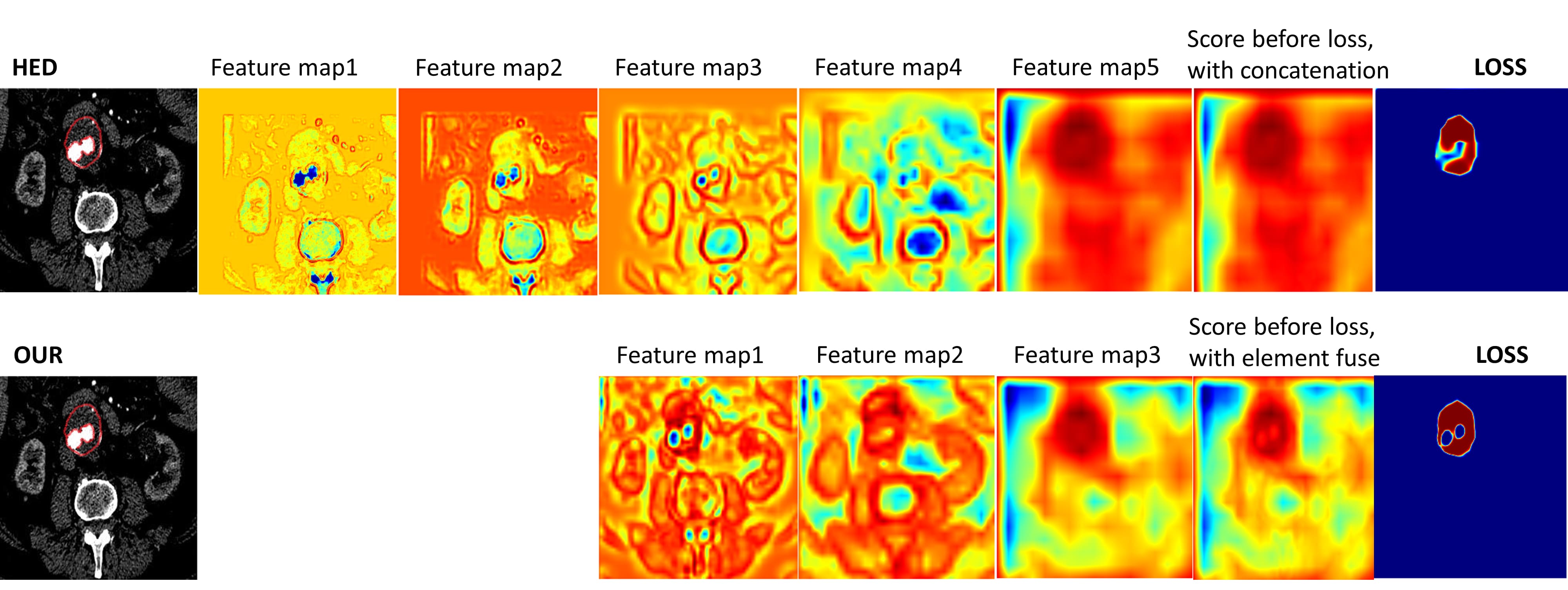}
\caption{\label{fig:hed_maps}Output side feature maps from original HED network and our proposed architecture, from shallower on the left to deepest and fusing of the maps on the right.}
\end{figure}

Secondly, the original HED network has been fine tuned to minimize a softmax loss. The goal is to get more precision when detecting edges than with the FCN networks, to better differentiate the thrombus from adjacent structures. The feature side maps in the HED net are extracted from the very beginning of the network and from the output of the last convolutional layer in each stage, instead of after the pooling, which reduces the need of large deconvolutions and increases the resolution of the output prediction. After training, the responses of the output side feature maps before concatenation revealed that, indeed, the shallower connections provide very rich edge information while the final maps are more coarse and fuzzy (see Figure~\ref{fig:hed_maps}). However, these first richer maps fail to detect the thrombus borders, while responding strongly to vertebrae, lumen and kidney borders that appear more contrasted. On the other hand, the deepest map seems to correctly locate the thrombus area and has the highest influence after concatenation, as expected from the performance of FCN46, which lead us to propose this new architecture. \par

The rationale for our new network architecture definition relies on the assumption that a combination between fine edge detection to distinguish the thrombus from adjacent structures (HED contribution) and global appearance and shape information (FCN46 contribution) is required for a correct segmentation. Hence, we have modified the original HED network by:

\begin{itemize}
\item Removing the first two output side connections, since their response is higher for other structures whose borders are more contrasted than for the thrombus edges. Deeper coarse fuzzy maps produce global information that is very valuable for thrombus segmentation, as seen in FCN46
\item Removing the initial padding to improve resolution
\item Specifying a crop offset of 1 pixel to avoid removing important image areas and to have a resultant centered image, which was not considered in the original HED network (see Figure ~\ref{fig:hed_maps})
\item Element wise fusing instead of concatenation to keep strongest activations 
\end{itemize}

\begin{figure}[htb]
\centering
\includegraphics[width=0.9\textwidth]{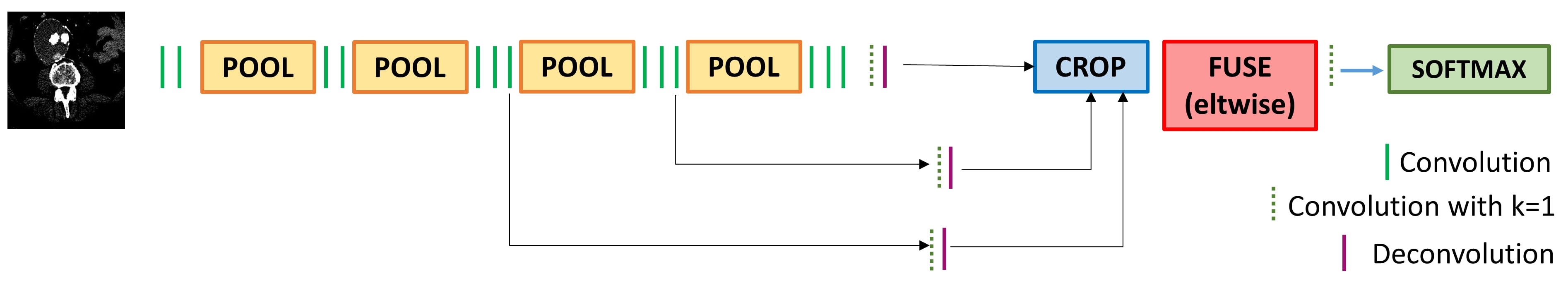}
\caption{\label{fig:hed_modif}Simplified scheme of the proposed modified HED network for thrombus segmentation.}
\end{figure}

\begin{figure}[htb]
\centering
\includegraphics[width=0.9\textwidth]{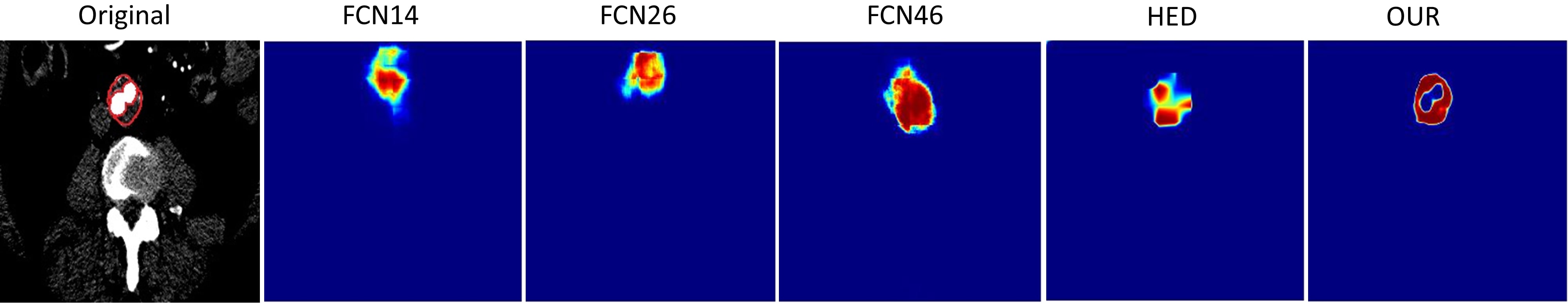}
\caption{\label{fig:loss}Probability maps obtained with FCN, HED and our network architecture for a sample image.}
\end{figure}

Among these modifications, the removal of the side connections and the substitution of the concatenation for the element wise fusing have the most relevant impact in the improved performance of the network. Figure \ref{fig:hed_maps} shows feature maps at different scales for both the original HED and our proposed network, as well as the final fusing of the maps using concatenation and element-wise fusing, respectively, which clearly depicts the improvements obtained with our adaptations. Figure \ref{fig:hed_modif} presents the simplified scheme of our proposed network architecture.
This network is optimized to produce very accurate thrombus segmentation with low computational and memory requirements. A comparison among the probability maps obtained with FCN, HED and our network architecture for an example image can be observed in Figure \ref{fig:loss}. \par  

As previously explained, we have applied cross-validation, meaning that each network architecture is trained and tested 4 times, each of them with a different training and testing data subset. Thus, we create four network instances for each architecture and average the results to provide more robustness to our approximation. The parameters employed for training the networks are summarized in Table~\ref{tab:trainingParams}. The function to be minimized in all the segmentation networks is the multinomial logistic loss for a one-of-many classification task, passing predictions through a softmax to get a probability distribution over classes: 

\[L = \frac{1}{N} \sum_{i=1}^{N}-log(\frac{e^{f_{y_{i}}}}{\sum_{j}e^{f_j}}),\]

where \(f_j\) denotes the \textit{j}-th element (\textit{j} $\in [1,k]$, \textit{k} being the number of classes) of the vector of class scores \textit{f}, and \textit{N} is the number of training data.
In all the cases we use a step-wise learning rate decay, with gamma equal to 0.1, and we train during 100 epochs. The FCN networks are fine-tuned from weights of the original FCN networks; the HED and our network are fine-tuned from the weights from the original HED. Data preprocessing and augmentation are not used in any case.

\begin{table}[htb]
\centering
\begin{tabular}{c|c|c|c|c|c|c}
\hline
Net & Optimizer & LR & Momentum & \thead{Weight \\ decay }& \thead{Batch \\ size} & \thead{Batch \\ accumulation} \\
\hline
FCN14 & SGD &  1e-3 & 0.9 & 1e-05 & 2 & 2\\
FCN26 & SGD & 1e-3 & 0.9 & 1e-05 & 2 & 2\\
FCN46 & SGD & 1e-3 & 0.9 & 1e-05 & 2 & 2\\
HED & SGD & 1e-3 & 0.9 & 1e-05 & 4 & -\\
OUR & SGD & 1e-3 & 0.9 & 1e-05 & 4 & -
\end{tabular}
\caption{\label{tab:trainingParams}Training parameters for the segmentation networks, where SGD refers to the Stochastic Gradient Descent optimization method and LR to the Learning Rate.}
\end{table}



\subsubsection*{3D binary mask generation} 
The outputs provided by the networks are 2D probability maps, where each intensity value is the probability of that pixel being thrombus or not. Thus, an automatic binarization of these maps is included as the last step. First, we reconstruct a 3D prediction map volume and apply a Gaussian filter in the z-direction. Doing so, we pretend to avoid sudden appearance and disappearance of anatomical structures and to ensure a continuity in the z-direction of our segmentation, improving the limitations of the 2D approximation. We set the sigma value to \(\sigma = 2*Spacing_{z}\). Then, K-means clustering of the 3D probability map is applied, where the number of clusters is fixed to 6, experimentally. The output clustered image is binarized, by removing the class with the lowest probability of being thrombus. A subsequent connected component analysis is used to keep the largest object, i.e. the thrombus.

\subsection{3D quantitative evaluation}
A comparison between the automatically obtained 3D binary segmentation masks and the manually delimited volume is necessary to evaluate the segmentation quality. For that purpose, we compute total overlap, Jaccard coefficient, Dice coefficient, false negative rate (FN) and false positive rate (FP) between the automatically segmented volume (source, S) and the ground truth (target, T), as proposed in ~\citep{itk}. \newline \par
\(Total\ overlap\ for\ thrombus\ region\ (r):\ \mid S_{r} \cap T_{r} \mid / \mid T_{r} \mid\) \par
\(Jaccard\ coefficient\ for\ thrombus\ region\ (r):\ 2 \mid S_{r} \cap T_{r} \mid / \mid S_{r} \cap T_{r} \mid\) \par
\(Dice\ coefficient\ for\ thrombus\ region\ (r):\ 2 \mid S_{r} \cap T_{r} \mid / (\mid S_{r} \mid + \mid T_{r} \mid) \) \par
\(False\ negative\ error\ for\  thrombus\ region\ (r):\ \mid T_{r}/S_{r} \mid / \mid T_{r} \mid\)\par
\(False\ positive\ error\ for\ thrombus\ region\ (r):\ \mid S_{r}/T_{r} \mid / \mid S_{r} \mid\)

\section{Experiments and results}
\label{sec:results}

\subsection{Thrombus region of interest detection}

The proposed thrombus region of interest detector, based on the \textit{DetectNet} architecture, aims at detecting the presence or absence of the thrombus in every 2D slice of a CTA volume and to determine a 2D bounding box around it, generating a region of interest appropriate for the upcoming segmentation. Hence, there are 2 objectives: 1) detect the initial and final slices (Zmin and Zmax) where the thrombus is visible from the whole 3D volume, and 2) determine a rectangular region around the thrombus in each 2D slice. \par

The determination of the initial and final slice where the thrombus appears is a complex task.  From a clinical point of view, theoretically, when the aortic wall expands more than 5 mm, then the presence of thrombus can be assumed. However, in practice, the delineation of the region of interest is done approximately by visual inspection, therefore inter- and intra-observer variability exists. 
To test the processed results of the network regarding the presence or absence of the thrombus in each slice of the CTA volume, we have compared the automatically selected slice range with the span selected by 3 different experts. The inter-observer variability is noticeable, where the manually indicated initial and final thrombus slices may differ depending on the observer's judgment. Hence, the inter observer variability (IOV) is measured for the initial and last slices of the thrombus, as mean and standard deviation of the selections of the three observers. Our interest relies on selecting a reduced region from the whole CTA that includes the thrombus, even if it includes also some adjacent slices. Hence, we concentrate on minimizing the false negative rate (FNR), defined as the ratio between the undetected thrombus slices and the total thrombus slices considering the mean of all the expert observers. The mean FNR for all the datasets and networks is 0.085\(\pm\)0.108. The results are summarized in 
Table~\ref{tab:detect fn}. The largest false negative rates of M1 and M3 are due to datasets with special characteristics: one of them is a very large thrombus comparing to the mean size, which also affects the segmentation as will later be explained; in the other one, the thrombus expands to the iliac arteries, which the observers have considered as part of the infrarenal thrombus, but the network discards it since this is not considered in the training.  

\begin{table}[htbp]
\centering
\begin{tabular}{c|c|c|c}
\hline
{} & {IOV initial slice} & {IOV final slice} & {FNR automatic detection}\\
\hline
{M0} & {3.6667\(\pm\)1.5275} &  {1\(\pm\)1} & {0.0087\(\pm\)0.0103} \\
{M1} & {13\(\pm\)12.49} &  {1.6667\(\pm\)1.1547} & {0.0833\(\pm\)0.0744} \\
{M2} & {24\(\pm\)5.1961} &  {6.3333\(\pm\)2.887} & {0.0523\(\pm\)0.0454} \\
{M3} & {5\(\pm\)2.6458} &  {9.3333\(\pm\)7.0556} & {0.194\(\pm\)0.1683} \\
\end{tabular}
\caption{\label{tab:detect fn}Inter observer variability (IOV) when selecting the initial and final thrombus slices and false negative rate (FNR) computed as the ratio between the undetected thrombus slices and the total thrombus slices considering the mean of all the expert observers for each instance.}
\end{table}

\begin{figure}[htb]
\centering
\includegraphics[width=1\textwidth]{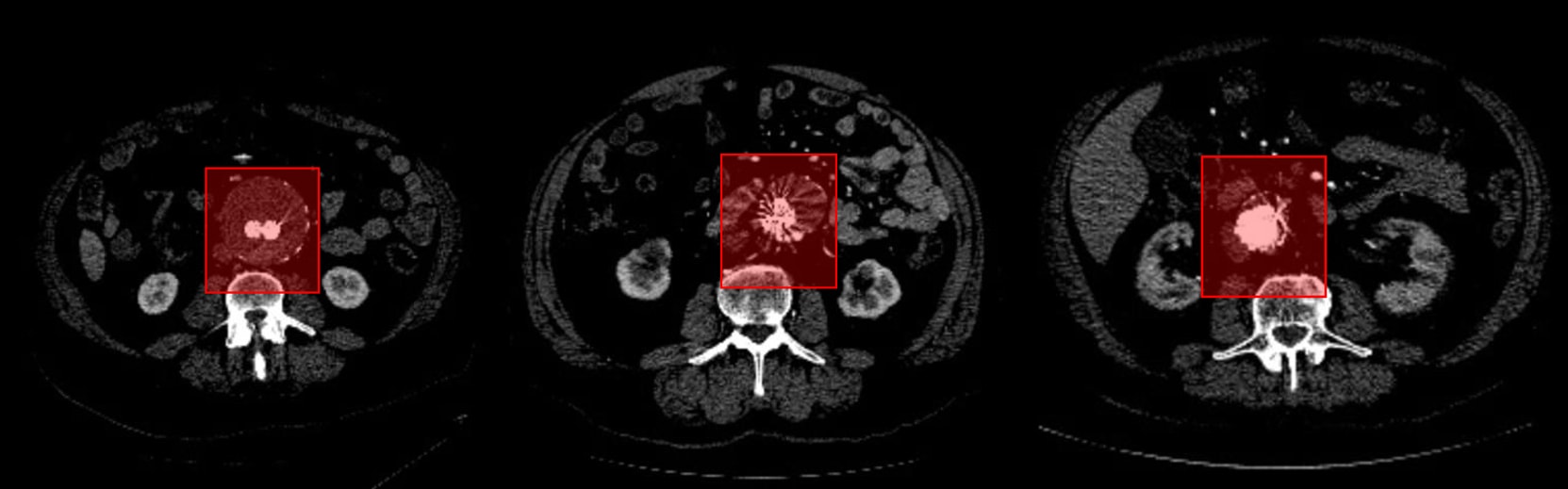}
\caption{\label{fig:det_res} Examples of some detected bounding boxes in the three datasets.}
\end{figure}

Regarding the 2D bounding boxes delimiting the thrombus in each slice, the minimum and maximum \textit{x} and \textit{y} coordinates of all the bounding boxes are selected. We then expand this region to include wider contextual information, which is necessary for a good segmentation. Thus, we always obtain a 3D region of interest that correctly delimits the thrombus in \textit{x} and \textit{y} in all the cases, for which some visual examples are presented in Figure~\ref{fig:det_res}.

\subsection{Thrombus segmentation}
Hereby, we present results of our novel network architecture that detects edges but preserves the global appearance of the thrombus, inspired both by FCN and HED networks. For comparison purposes and to confirm that our proposed architecture indeed improved the final segmentation accuracy, effectively benefiting from both FCN and HED network properties, we have fine-tuned and tested each of these networks and compared the results with our approach. As explained in the methods section, each of the architectures is trained and tested with a 4-fold cross-validation approach, generating 4 instances per network architecture and the mean of the quantification results is computed.\par 

First, we have fine-tuned the FCN14, FCN26 and FCN46 networks as explained in Sec.~\ref{sec:methods}. 


We have then tested the four network instances of these three FCN architectures with the set generated from previously unseen images of the remaining three patients not used for the training in each cross-validation instance. After running the post-processing algorithm to get the 3D binary segmentation, the quantitative comparison with the binary ground truths delimited by our clinical expert is accomplished. Results are shown in Table~\ref{tab:results}, as mean and standard deviation of all testing datasets for the 4 cross-validation instances. As expected, the FCN46 network yields better results, since it preserves the smooth shape of the thrombus, although it fails to differ it from some adjacent structures, hence over segmenting it. \par


Secondly, the HED network and our proposed architecture have been fine-tuned and compared, to prove that the changes applied to the HED net, inspired by FCN46, indeed notably improved the segmentation of the thrombus. Table~\ref{tab:Comptraining} compares the FCN networks with the HED and our proposed network in terms of training and validation loss, training and testing time, learned parameters and memory. The provided values refer to the means of all the instances of each network architecture. With our approach, the number of learned model parameters (i.e. weights and biases) are reduced, the training and validation losses decrease notably and the training and testing times are also inferior compared to the other networks.

\begin{table}[htb]
\centering
\begin{tabular}{c|c|c|c|c|c|c}
\hline
Net  &  \thead{Training \\ loss} &  \thead{Validation \\ loss} & \thead{Learned \\ parameters} & \thead{Memory \\ (MB)} & \thead{Training \\ time} & \thead{Testing \\ time} \\
\hline 
FCN14 & 0,017 & 0,024 & 134.409,791 & 512 & 3h52 & 26 image/s\\
FCN26 & 0,009 & 0,016 & 134.486,882 & 513 & 3h50 & 26 image/s\\
FCN46 & 0.013 & 0,011 & 134.702,069 & 513 & 4h01 & 27 image/s\\
HED  & 0,010 & 0,018  & 14.717,541 & 56 & 4h34 & 29 image/s\\
OUR & 0,004  & 0,009 & 14.717,322 & 56 & 2h39 & 35 image/s
\end{tabular}
\caption{\label{tab:Comptraining}Comparison between original HED training and modified HED training.}
\end{table}

\begin{figure}[htb]
\centering
\includegraphics[width=0.8\textwidth]{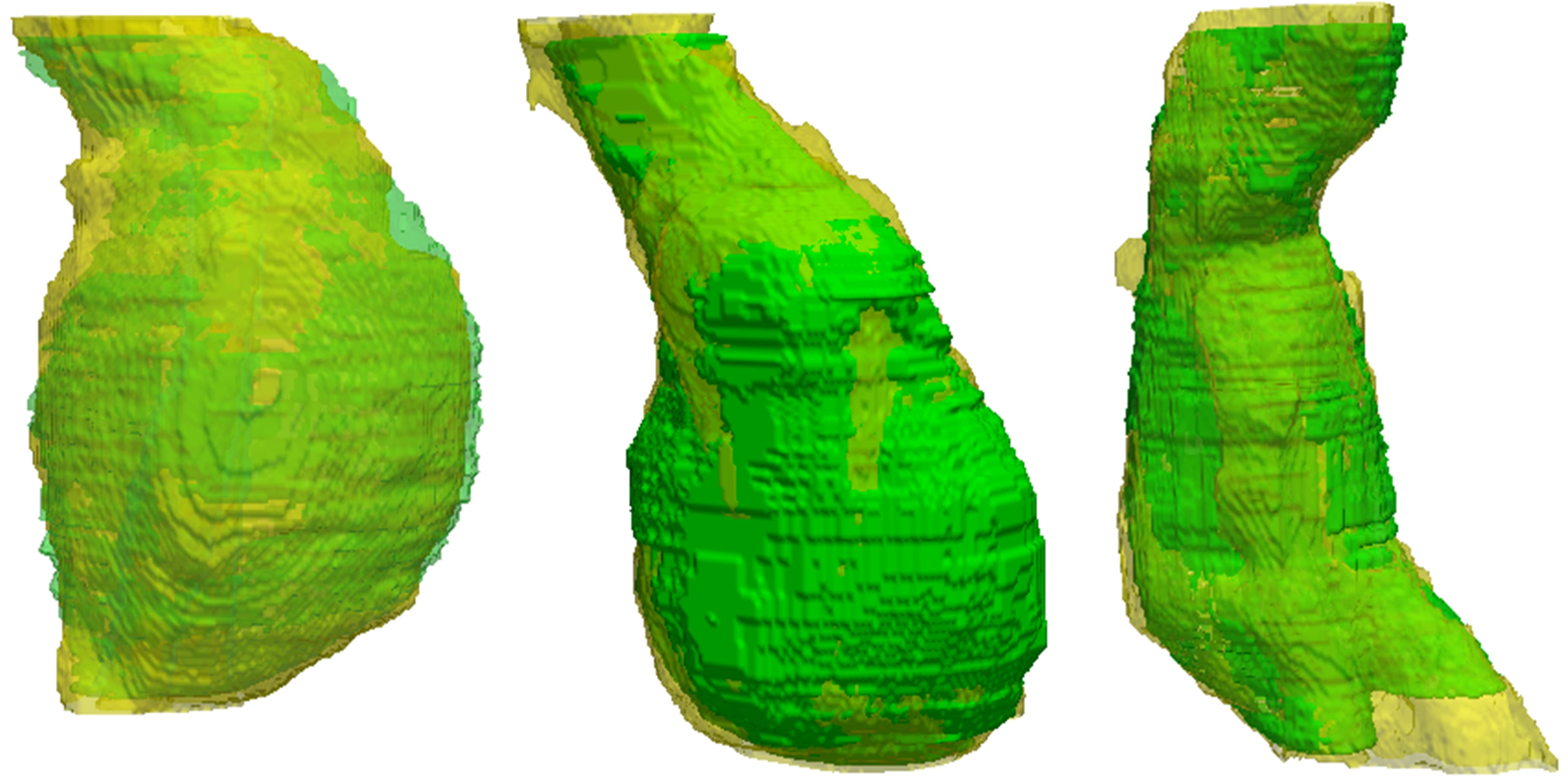}
\caption{\label{fig:qua_res}3D Qualitative results of the obtained segmentation for three example datasets. Ground truth are represented in green, while automatic segmentations are display in yellow.}
\end{figure}

After testing both HED and our network, the 3D binary segmentation masks are generated. A qualitative evaluation of 3 example dataset segmentations obtained with the proposed network is shown in Figure~\ref{fig:qua_res}.

\begin{figure}[htb]
\centering
\includegraphics[width=0.8\textwidth]{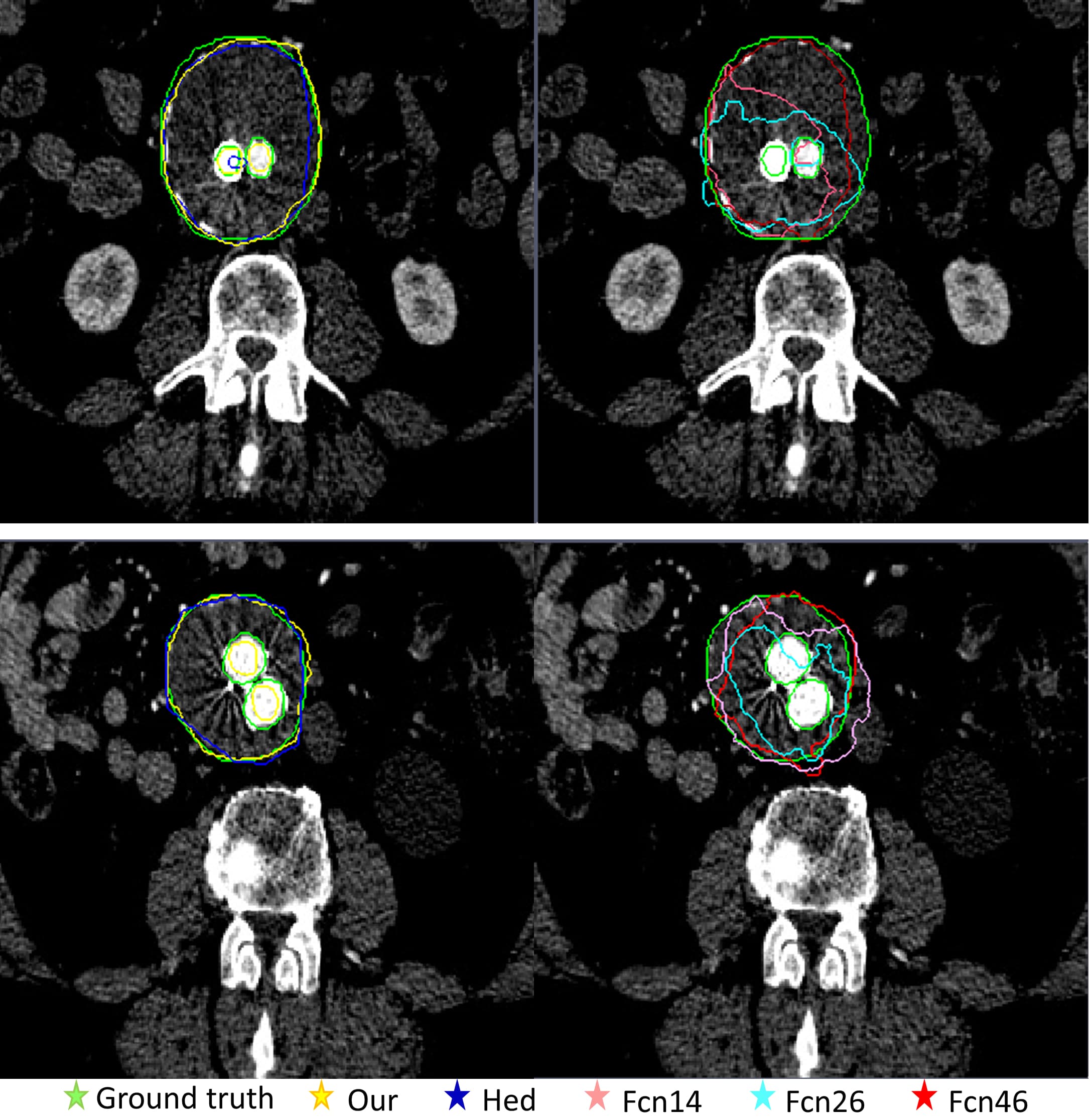}
\caption{\label{fig:res2D}2D qualitative results obtained with the different networks, where thrombus contours are outlined for two different datasets, each in one row.}
\end{figure}

Figure~\ref{fig:res2D} presents example results in 2D for two different datasets (one per row), where the outline of the ground truth and the automatically obtained segmentation is superimposed onto the original image for all the network architectures.  \par

Total overlap, Jaccard coefficient, Dice coefficient, FN and FP are computed to evaluate the results of the automatic DCNN-based segmentation against the ground truth segmentations. This quantitative evaluation is summarized in Table~\ref{tab:results} as the mean values and standard deviations of the four instances created for each network architecture. Testing results show an increase in the Dice coefficient of about 13.9~\% with our architecture, comparing to the HED network performance. The false positive rate is reduced almost by half the value of the HED network and we also improve the false negative rate by 7~\%. We have observed that the segmentation quality for one dataset is notably inferior to the mean in all the network architectures since it is always sub-segmented. For our proposed architecture, the total overlap for this dataset is 61.8~\%, which is much lower than the mean. The thrombus represents an endotension case (i.e. unfavorable AAA case where there is no visible leak but the thrombus size increases) and its mean diameter is larger than the diameter of the other datasets. However, utilizing low-scale features for edge detection that would not rely on the size of the thrombus but on the limits between thrombus and surroundings does not improve the segmentation results. Our hypothesis is that an endotension case could present a different texture content comparing to other thrombi, and thus the network may be unable to generalize for this dataset, but should be further investigated. This is promising since it could set ground for a hidden endoleak detector in the future.

\begin{table}[htb]
\centering
\begin{tabular}{c|c|c|c|c|c}
\hline
Net & \thead{Total \\ overlap} & Jaccard & Dice & FN & FP \\
\hline
FCN14 & 0.64\(\pm\)0.15 & 0.37\(\pm\)0.10 & 0.53\(\pm\)0.11 & 0.36\(\pm\)0.15 & 0.51\(\pm\)0.16 \\
FCN26 & 0.54\(\pm\)0.12 & 0.35\(\pm\)0.08 &  0.51\(\pm\)0.09 & 0.46\(\pm\)0.12 & 0.47\(\pm\)0.15\\
FCN46 & 0.75\(\pm\)0.16 &  0.44\(\pm\)0.12 & 0.60\(\pm\)0.11 & 0.25\(\pm\)0.16 & 0.47\(\pm\)0.15\\
HED & 0.83\(\pm\)0.08 & 0.56\(\pm\)0.10 & 0.72\(\pm\)0.09 & 0.17\(\pm\)0.08 & 0.36\(\pm\)0.13  \\
OUR & 0.84\(\pm\)0.11 & 0.70\(\pm\)0.10 & 0.82\(\pm\)0.07 & 0.16\(\pm\)0.11 &  0.18\(\pm\)0.08
\end{tabular}
\caption{\label{tab:results}Segmentation results for FCN, HED and our proposed network architecture. Provided values correspond to the mean of all cross-validation instances of each architecture.}
\end{table}

\section{Discussion and future work}
\label{sec:disc}
We have proposed a DCNN-based fully automatic segmentation of the abdominal aortic thrombus. From a complete post-operative CTA dataset, our method first detects the thrombus region within the whole volume. We have translated the computer vision \textit{DetectNet} architecture to the medical imaging domain and, to the best of our knowledge, this is the first work in which object detection in medical imaging is done from the whole 2D slice, instead of patch-wise. Thrombus segmentation is then achieved from the previously localized region, reducing the amount of memory and computation time needed to obtain an accurate segmentation. We have validated the segmentation by comparing the 3D automatic segmentation with a ground truth delimited by a clinical expert and we have provided a quantitative evaluation of the results. 

As presented in Sec.~\ref{sec:soa}, previous approaches to thrombus segmentation required additional input both in the form of lumen segmentation and centerline extraction, and user interaction to control the segmentation, setup the necessary parameters or edit the obtained segmentation. Thus, the methods were not automatic and robust enough for their usage in the clinical practice, and much slower than our approach. In addition, some of the previous methods relied on a certain thrombus shape prior to control the segmentation process, which is difficult to model due to the inherent irregular structure of the thrombus surface. Our method is fully automatic, does not require neither user interaction in any of the steps nor a priori knowledge of the shape of the thrombus. 

The obtained results exceed thrombus state-of-the-art segmentation, being the mean total overlap and Dice score between the segmentation and the ground truth around 84\% and 82\%, respectively. Training and testing datasets are CTA volumes obtained with different scanner manufacturers and models, all of them extracted from a real clinical environment. The patient is always located in supine position and CTA starts around the diaphragm and expands to the iliac crest. The presence of stent metal artifacts is limited and the noise is variable, but always low. The employed datasets have also a varying spatial resolution, which confirms the robustness of the method although a larger number of datasets, some of them rotated, should be tested in the future.
We have included a cross-validation approximation, creating 4 network instances per architecture that are trained and tested with a different data subset to improve robustness. Then, the 3D binary segmentation is extracted from the slice-by-slice prediction maps with a naive K-means processing approach. The ground truth for the datasets has been delineated manually by our clinical expert. To evaluate the benefit of the network in terms of time consumption, 3 datasets were segmented manually slice by slice, which took our expert about three hours. With our approach, a dataset segmentation can be obtained approximately in a minute, including the binary mask segmentation step. 
\par

During our experiments, we have proved that combining features from the edge detection neural network with network architectures that extract more global appearance information notably improves the segmentation of anatomical structures such as the thrombus. The proposed architecture combines global characteristics as in FCN46 network, where there was no skip connection, with the local and fine edge segmentation approach proposed by the HED network. Connecting the output of the convolutional layers instead of the pooling layer as proposed in HED improves the resolution of the segmentation, since large deconvolutions are avoided. By removing the initial skip connections in the HED network we have been able to better preserve the thrombus shape while distinguishing between the thrombus and adjacent structures of similar intensity. The proposed architecture is faster than both FCN and HED networks, requires less memory, less parameters and the training, validation and testing accuracy are higher. We have trained and tested the networks in 2D slice-by-slice, which reduces the need of a large number of 3D datasets and provided us with the ability of utilizing pretrained networks. \par

To evaluate the clinical applicability of the obtained segmentation results for disease progression and rupture risk assessment during follow up, we have computed the relative volume difference between the segmentations obtained with our proposed architecture and the corresponding ground truths. 
Intra-observer and inter-observer variability for volume measurements from semi-automatically segmented aneurysms can go up to 5\% ~\citep{Cha02,parr11,van08}. From a clinical perspective, our pipeline produces significant volume differences between the ground truth and the automatic segmentation, being the mean volume difference of 11.6\%. Three out of 13 datasets show very large discrepancies between the volume of the automatic segmentation and the ground truth's volume: the first one presents a very large thrombus compared to the others, as previously mentioned, and is notably subsegmented; another dataset has a thrombus that expands to the iliac artery, and since the network was not trained sufficiently for these cases that area is not correctly segmented, again showing a large relative volume difference as compared to the ground truth; the last dataset corresponds to the dataset with the smallest thrombus, for which a minimal segmentation error has a larger effect in the quantified volume. If these three datasets are removed, a mean volume difference of 4.11\% is obtained, which lays below the maximum variability of human observers. Due to the large variability of the data employed in this work and considering that we did not discard patients with outlying characteristics, we consider that the results can be extrapolated to future cases, since we assume that by training the networks with more datasets with these characteristics the mean relative volume difference would lay within the experienced human observer variance without the need of human intervention. 

As compared to the literature, only two studies that evaluate the relative volume difference have been found. In ~\citep{Mot10} an automatic thrombus segmentation approach, based on a previous lumen segmentation and tested with 8 CTAs, achieved relative volume differences of 8\% with a mean volumetric overlap error of around 13\%. Another study presented in ~\citep{Zoh12} reports volume differences of 4\% and a total overlap error of 6\%, but segmenting the thrombus semi-automatically. Thus, our results are comparable to the state of the art percentages but with a fully-automatic segmentation approach that does not require any additional input apart from the CTA itself and that is evaluated against more datasets with different characteristics. In relation to the applicability of the proposed pipeline for disease evolution assessment during follow-up, the results still need to be refined. EVAR reporting standards ~\citep{Cha02} state that an increase in the aneurysm volume of 5\% is considered clinically relevant and a clinical failure after EVAR. With our approach, over-segmentation and sub-segmentation occur equally, and thus, even if the calculated mean relative volume difference lays within the expert's inter-observer rate, it cannot be directly translated to the clinical practice. For that purpose, the network should always over-segment or sub-segment the thrombus, so that following always the same pattern a comparison between two time points would be feasible, but no study has been found in literature that deals with this problem. In spite of that, we provide the clinician with a good initial approximation that could be further refined manually in a notably inferior time as doing it manually from the beginning, which is the major problem for which volume based evolution assessment is still not performed in the clinical practice. 


Regarding future work, we aim at reducing the volume difference between ground truth and automatically segmented thrombus, by adapting our method to that purpose and analyzing the volume quantification results with more data. Additionally, a network trained not only with postoperative datasets but also with preoperative data is being investigated. Besides, multiclass segmentation for simultaneous lumen and thrombus extraction is an interesting line of research. Lumen segmentation in the post-operative scenario is challenging due to contrast inhomogeneity and appearance of artifacts due to the presence of the stent-graft. DCNN-based lumen segmentation could solve some of the issues traditional intensity-based approaches present.  

\section*{Acknowledgements}
This work has been supported by the DEFENSE (TIN2013-47913-C3) research project, funded by the Ministry of Economy and Competitiveness-Government of Spain.

We also gratefully acknowledge the support of NVIDIA Corporation with the donation of the Titan X GPU used for this research.

\bibliographystyle{model2-names}
\bibliography{Biblio}

\end{document}